\documentclass[conference]{IEEEtran}
\usepackage{graphicx}

\ifCLASSINFOpdf
\else
\fi
\begin{document}
%
\title{ICDAR 2015 Text Reading in the Wild Competition}

\author{\IEEEauthorblockN{Xinyu Zhou, Shuchang Zhou, Cong Yao, Zhimin Cao, Qi Yin}
\IEEEauthorblockA{Megvii Inc.\\
Beijing, 100190, China\\
Email: \{zxy, zsc, yaocong, czm, yq\}@megvii.com}}


%


\maketitle

\begin{abstract}
Recently, text detection and recognition in natural scenes are becoming increasing popular in the computer vision community as well as the document analysis community. However, majority of the existing ideas, algorithms and systems are specifically designed for English. This technical report presents the final results of the ICDAR 2015 Text Reading in the Wild (TRW 2015) competition, which aims at establishing a benchmark for assessing detection and recognition algorithms devised for both Chinese and English scripts and providing a playground for researchers from the community. In this article, we describe in detail the dataset, tasks, evaluation protocols and participants of this competition, and report the performance of the participating methods. Moreover, promising directions for future research are discussed.
\end{abstract}


%
\IEEEpeerreviewmaketitle

\section{Introduction}

Due to the practical utility and ubiquity of scene text, text detection and recognition in natural scenes have become important, active research topics in both the computer vision community and the document analysis community. This trend is evidently confirmed by the dramatic increase of related research papers~\cite{Ref:Epshtein2010, Ref:Neumann2010, Ref:Wang2011, Ref:Yao2012, Ref:Yao2014, Ref:Yao2014C, Ref:Jaderberg2014, Ref:Zhu2015} in recent years. Considerable progresses and obvious improvements have been achieved, mainly driven by the competitions and public datasets in this area, such as the ICDAR Rubust Reading competitions~\cite{Ref:Lucas2003, Ref:Lucas2005, Ref:Shahab2011, Ref:Karatzas2013}, MSRA-TD500~\cite{Ref:Yao2012}, SVT~\cite{Ref:Wang2011}, Chars74K~\cite{Ref:deCampos2009} and IIIT-5K Word~\cite{Ref:Mishra2012B}.

However, upon close observation and investigation, we found that most of the previous systems and datasets fall short in at least two aspects: (1) Though there are more than 100 kinds of languages that are widely used all over the world, majority of these algorithms can only handle texts of English (or other Latin-root languages). How well they could perform on texts of other languages (for instance, Chinese, Kannada, Thai and Hebrew) is unclear. (2) The diversity and difficulty of the existing datasets do not match real-world complexity in real applications, because the sizes and image sources of these datasets are limited. To break through these limitations, methods that can deal with multilingual texts in the wild are desirable. Accordingly, datasets containing multilingual texts with real-world complexity and corresponding evaluation protocols are essential prerequisites.

Therefore, we organized the ICDAR 2015 Text Reading in the Wild (TRW 2015) competition\footnote{http://icdar2015.imageplusplus.com/}, which generates a large-scale text image database, proposes two text detection or recognition tasks and devises corresponding evaluation methods. This competition can serve as a standard benchmark for assessing algorithms that are designed for multilingual text detection and recognition in complex natural scenes. To the best of our knowledge, the dataset in this competition is the first that can be used for evaluating detection and recognition algorithms for both Chinese and English scripts.

One thing worthy mentioning is that this competition is just a starting point and the main goal is to provoke interest and enthusiasm from the community. We believe that more competitions, datasets and algorithms that involve multilingual text understanding in natural scenes will appear in the near future.

\section{The Competition}

\subsection{Dataset and Annotations}

The dataset of this competition includes about 1000 natural images, which are harvested from the Internet or taken by volunteers. 500 images are selected for algorithm development and validation, and 484 image for testing. For each image, the polygons and content of all the text lines within it are annotated. It is allowed to use extra data for training in this competition.

\begin{figure}[!tp]
\centering
\includegraphics[width=0.9\linewidth]{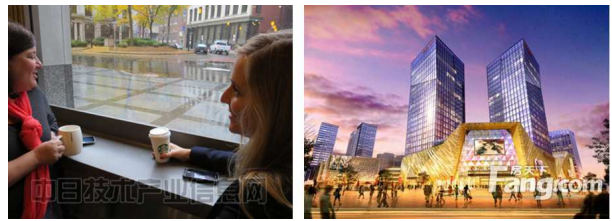}
\caption{Images with translucent characters.} \label{Fig:Translucent}
\vspace{-5mm}
\end{figure}

\begin{figure}[!tp]
\centering
\includegraphics[width=0.9\linewidth]{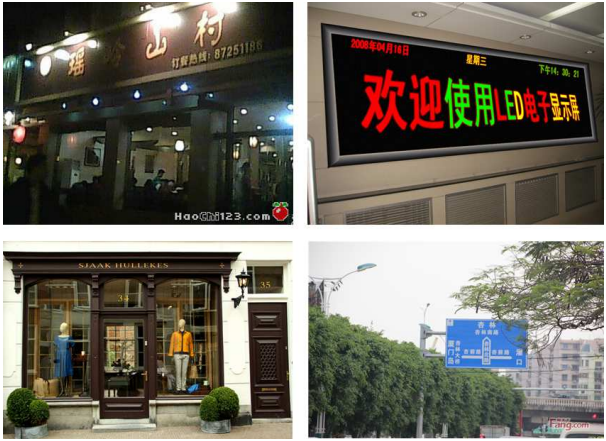}
\caption{Images with Chinese or English scripts.} \label{Fig:Other}
\vspace{-4mm}
\end{figure}

The text lines may be divided into one of the four categories: (1) Translucent English; (2) Translucent Other; (3) Non-Translucent English; (4) Non-Translucent Other. The categories with "Translucent" indicate presence of translucent text, which may be used to encode website link, name of shop, contact information, etc.. Reading the encoded text will help determine if the text is in line with anti-spam policy of the site hosting the images. Samples of such images are shown in Fig.~\ref{Fig:Translucent}. The categories with "Other" indicate the presence of multilingual text comprising of Chinese and English in natural/Internet images. Several examples are depicted in Fig.~\ref{Fig:Other}.

As can be seen, the dataset is both diverse and challenging, since the images are real-world natural images from different sources and almost all the images are taken or generated by non-professionals.

\subsection{Tasks and Evaluation Protocols}

There are two tasks in this competition: \textbf{Text Locating} and \textbf{Text Recognition}. For text locating, given an input image, you should produce a set of polygons in the image, which will be deemed as text line candidates. For simplicity, we adopt the evaluation method from the ICDAR 2003 Robust Reading competition, with the only difference being that we use polygon intersection area rather than rectangle in evaluation. For text recognition, given an image containing a single word, clause or sentence, you should output a sequence of characters denoting the textual content in that image. We evaluate the performance of algorithms by case-sensitive normalized edit distance.

In the training data, we provide coordinates of the text lines. Participants only interested in cropped image recognition are free to crop the text line images with help of these coordinates, as long as they do not use additional human annotations in this process. For example, one can take a $15\%$ larger text line image expanded from the given coordinates.

\subsection{Participating Methods}

There were several teams registered the competition, but only two teams submitted valid results before the deadline. The Stradvision\footnote{Hojin Cho, Myungchul Sung, and Bongjin Jun. StradVision, Inc., Korean.} team participated in the Text Locating task while the CASIA\_NLPR\footnote{Yi-Chao Wu, Xin He, Zhuo Chen, Kai Chen, Fei Yin, and Cheng-Lin Liu. National Laboratory of Pattern Recognition, Institute of Automation of the Chinese Academy of Sciences, Beijing, China} team participated in the Text Recognition task. The brief descriptions of these methods are as follows:

\subsubsection{Stradvision}

First, we extract character candidates using extremal regions (ER). Then, we verify the extracted character candidates with the character classifier trained by Agile Learning\footnote{http://www.stradvision.com/}. Afterwards, we do text-patch matching which greatly enhances the recall rate, and group the characters into text regions.

\subsubsection{CASIA\_NLPR}

For the text extraction, we extract text connected components (CCs) in the YIQ color space. First we binarize an image into high-value and low-value CCs using OTSU’s algorithm in each channel. Then for each channel we select high-value or low-value CCs using an classifier with features characterizing the geometric relationship of the two sets of CCs. At last we select one channel as the text extraction result by comparing the numbers and areas of text CCs of all three channels after using a non-text/text CC classifier.

In text word recognition, the word image is first over-segmented into primitive segments using an MLP with 968-D features for candidate cut classification. Based on over-segmentation, the word image undergoes lexicon-free recognition with a statistical language model~\cite{Ref:Wang2012}. After text line recognition, we analyze the result to correct the case of letters and filter out some characters based on common sense.

\section{Results}

\subsection{Text Locating}

The performances of the algorithms participated in the Text Locating task are shown in Tab.~\ref{Tab:TextLocating}. The baseline method we adopted is an online service provided by an international IT enterprise. The Stradvision method significantly outperforms the baseline method (0.759 vs. 0.457 in F-Measure).

\begin{table}
\caption{Performances of algorithms participated in the Text Locating task.} 
\label{Tab:TextLocating}
\begin{center}
\begin{tabular}{|c|c|c|c|}
\hline
\textbf{{\scriptsize Algorithm}}&{\scriptsize Precision}&{\scriptsize Recall}&{\scriptsize F-Measure}  \\
\hline
\hline
{\scriptsize Stradvision}&{\scriptsize 0.787}&{\scriptsize 0.734}&{\scriptsize 0.759}  \\
\hline
{\scriptsize Baseline}&{\scriptsize 0.721}&{\scriptsize 0.335}&{\scriptsize 0.457}  \\
\hline
\end{tabular}
\end{center}
\vspace{-1mm}
\end{table}

\subsection{Text Recognition}

The performances of the algorithms participated in the Text Recognition task are shown in Tab.~\ref{Tab:TextRecognition}. The baseline method is the same online service as mentioned above. The CASIA\_NLPR method performs much better than the baseline method.

\begin{table}
\caption{Performances of algorithms participated in the Text Recognition task.} 
\label{Tab:TextRecognition}
\begin{center}
\begin{tabular}{|c|c|}
\hline
\textbf{{\scriptsize Algorithm}}&{\scriptsize Normalized Edit Distance}  \\
\hline
\hline
{\scriptsize CASIA\_NLPR}&{\scriptsize 0.279}  \\
\hline
{\scriptsize Baseline}&{\scriptsize 0.735}  \\
\hline
\end{tabular}
\end{center}
\vspace{-1mm}
\end{table}

Since there are no more entries that submitted legal results in time, we are not able to judge whether these submissions are the state-of-the-art on the dataset of this competition. However, from the numbers we can draw a rough conclusion that the participating methods, though achieved impressive performance, are far from meeting the requirements of real-world applications, just like previous algorithms in the literature~\cite{Ref:Epshtein2010, Ref:Neumann2010, Ref:Wang2011, Ref:Yao2012, Ref:Yao2014, Ref:Yao2014C}. There is still room for improvement in both text detection and recognition for Chinese and English scripts.

\section{Conclusion}

In this paper, we have presented the details of the ICDAR 2015 Text Reading in the Wild competition, including the dataset, tasks, evaluation protocols, participating methods and final results. As cam be seen, localizing and reading text in the wild, especially in multilingual scenarios (e.g. Chinese, English, Korean, etc.), are still extremely challenging tasks.

However, we believe that accurate and robust systems for multilingual text detection and recognition in natural scenes are on the point of realization, if the deep learning framework is utilized to make full use of the characteristics of scene text and background elements from large amount of data~\cite{Ref:Zhu2015}.






\bibliographystyle{IEEEtran}
\bibliography{TRW2015}

%




\end{document}